%% file: 0-article.tex
\documentclass{llncs}

\usepackage{url}
\usepackage{mdframed}
\usepackage{latexsym}
\usepackage{python}
\usepackage{wrapfig}
\usepackage{epstopdf}
\usepackage{array}
\usepackage{multirow}
\usepackage{caption}
\usepackage{wrapfig}
\usepackage{booktabs}

\usepackage{subfigure}
\usepackage{amssymb}
\usepackage{graphicx}
\usepackage{amssymb}
\usepackage{amsfonts}
\usepackage{bigstrut}

\usepackage[cmex10]{amsmath}
\usepackage{pgfplots}
\usepackage{algorithm}
\usepackage[noend]{algpseudocode}
\usepackage{footmisc}
\usepackage{rotating}
\usepackage{xparse}
\usepackage{todonotes}
\usepackage{comment}
\usepackage{enumitem}
\usepackage{marginnote}
\usepackage{xcolor}
\usepackage{bigstrut}
\usepackage{hyperref}
\usepackage{colortbl}
\usepackage{cleveref}

\input{addon/macros}

\begin{document}
\mainmatter
%
%\title{From Time Series to Process Model Forecasting}
\title{Process Model Forecasting Using Time Series Analysis of Event Sequence Data}
\titlerunning{Process Model Forecasting From Time Series } 
\author{Johannes~De~Smedt\inst{1}\orcidID{0000-0003-0389-0275}\and Anton~Yeshchenko\inst{2}\orcidID{0000-0002-5346-8358} \and Artem~Polyvyanyy\inst{3}\orcidID{0000-0002-7672-1643} \and Jochen~De~Weerdt\inst{1}\orcidID{0000-0001-6151-0504} \and Jan~Mendling\inst{4}\orcidID{0000-0002-7260-524X}}
\institute{
KU Leuven, Leuven, Belgium\\
\email{\{johannes.desmedt;jochen.deweerdt\}@kuleuven.be}\\
\and
Vienna University of Economics and Business, Austria\\
\email{\{anton.yeshchenko;jan.mendling\}@wu.ac.at}
\and
The University of Melbourne, Melbourne, Australia\\
\email{artem.polyvyanyy@unimelb.edu.au}
\and
Humboldt-Universit{\"a}t zu Berlin, Berlin, Germany\\
\email{jan.mendling@hu-berlin.de}
}
\maketitle

\begin{abstract}
Process analytics is an umbrella of data-driven techniques which includes making predictions for individual process instances or overall process models. At the instance level, various novel techniques have been recently devised, tackling next activity, remaining time, and outcome prediction. At the model level, there is a notable void. It is the ambition of this paper to fill this gap. To this end, we develop a 
technique to forecast the entire process model from historical event data. A forecasted model is a will-be process model representing a probable future state of the overall process. Such a forecast helps to investigate the consequences of drift and emerging bottlenecks. 
Our technique builds on a representation of event data as multiple time series, each capturing the evolution of a behavioural aspect of the process model, such that corresponding forecasting techniques can be applied.
Our implementation demonstrates the accuracy of our technique on real-world event log data.
\end{abstract}
\input{1-introduction}
\input{2-motivation}

\input{3-methodology}

\input{4-evaluation}
\input{5-conclusion}

\bibliographystyle{splncs03}
\bibliography{lib}

\end{document}

%% file: 1-introduction.tex
\section{Introduction}\label{sec:introduction}
Process analytics is an area of process mining~\cite{van2016data} which encompasses Predictive Process Monitoring (PPM) aimed at making predictions for individual process instances or overall process models. At the instance level, various novel PPM techniques have been recently devised, tackling problems such as next activity, remaining cycle time, or other outcome predictions~\cite{DBLP:conf/bpm/Francescomarino18}. These techniques make use of neural networks \cite{DBLP:conf/caise/TaxVRD17}, stochastic Petri nets \cite{DBLP:conf/icsoc/Rogge-SoltiW13}, and general classification techniques \cite{DBLP:journals/tkdd/TeinemaaDRM19}.

At the model level, there is a notable void. Many analytical tasks require not only an understanding of the current as-is, but also the anticipated will-be process model. A key challenge in this context is the consideration of evolution as processes are known to be subject to drift~\cite{maaradji2017detecting,DBLP:conf/bpm/PollPRRR18,yeshchenko2019comprehensive,yeshchenko2021visual}. A forecast can then inform the process analyst how the will-be process model might differ from the current as-is if no measures are taken, e.g., against emerging bottlenecks.

This paper presents the first technique to forecast whole process models. To this end, we develop a technique that builds on a representation of event data as multiple time series. Each of these time series captures the evolution of a behavioural aspect of the process model in the form of directly-follows relations (DFs), such that corresponding forecasting techniques can be applied for directly-follows graphs (DFGs). Our implementation on six real-life event logs demonstrates that forecasted models with medium-sized alphabets (10-30 activities) obtain below 15\% mean average percentage error in terms of conformance.
Furthermore, we introduce the Process Change Exploration (PCE) system which allows to visualise past and present models from event logs and compare them with forecasted models.

This paper is structured as follows. Section \ref{sec:2:motivation} discusses related work and motivates our work. Section \ref{sec:methodology} specifies our process model forecasting technique together with the PCE visualisation environment. Section \ref{sec:experiment} describes our evaluation, before Section \ref{sec:conclusion} concludes the paper.

%% file: 2-motivation.tex
\section{Related work and motivation}\label{sec:2:motivation}
Within the field of process mining, research on and use of predictive modelling techniques has attracted plenty of attention in the last five years. PPM techniques are usually developed with a specific purpose in mind, ranging from next activity prediction~\cite{evermann2017predicting,DBLP:conf/caise/TaxVRD17}, over remaining time prediction~\cite{verenich2019survey}, to outcome prediction~\cite{kratsch2020machine}. For a systematic literature review of the field, we refer to~\cite{neu2021systematic}. 
Beyond the PPM field, this work is related to previous research on stage-based process mining~\cite{nguyen2016business}, in which a technique is presented to decompose an event log into stages, and work on the detection of time granularity in event logs ~\cite{pourbafrani2020}. 

\begin{figure}
    \centering
    \includegraphics[width=\textwidth]{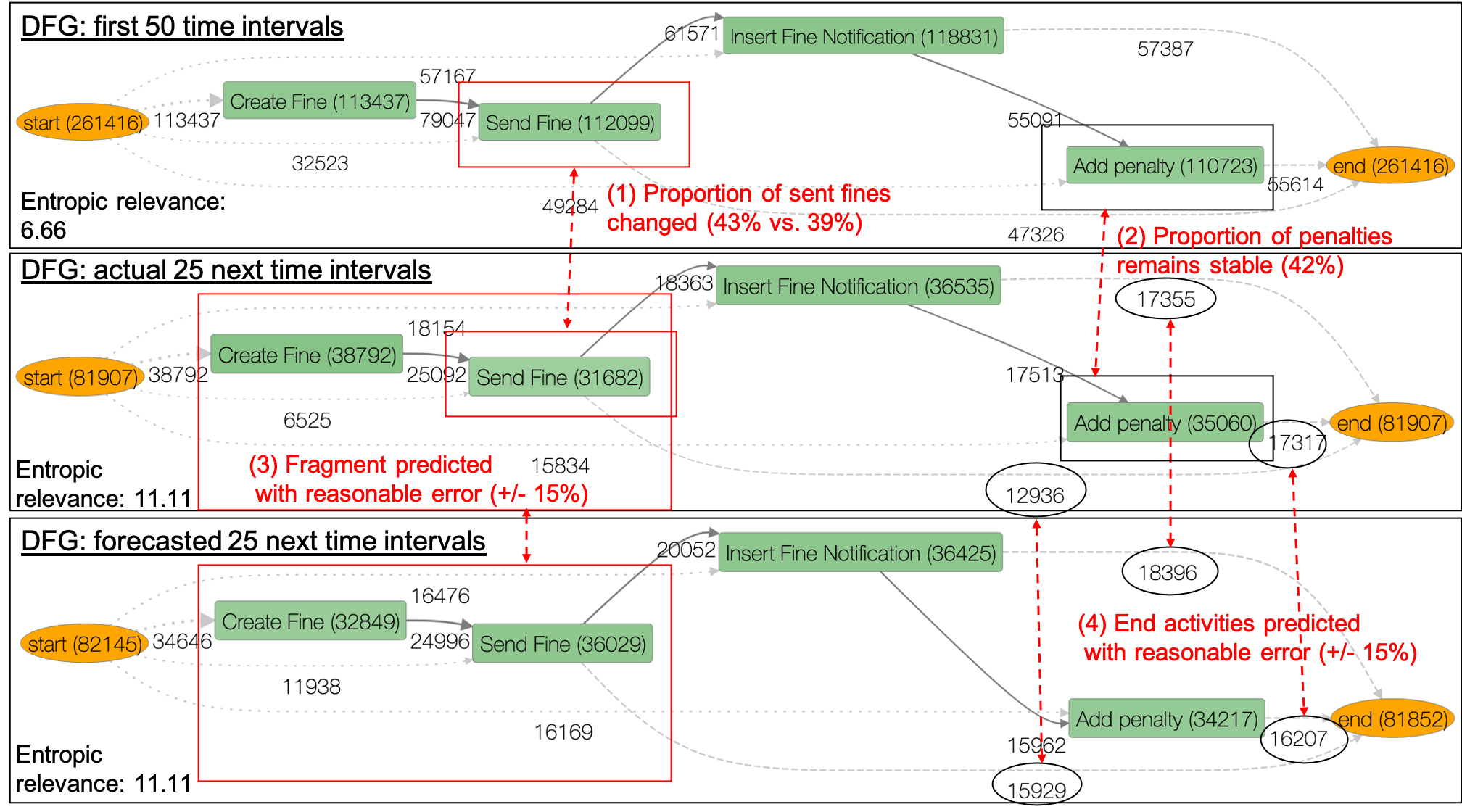}
    \caption{Directly-follows graphs of the 50 first intervals of the event log, as well as a forecasted and actual DFG of the 25 next intervals.}
    \label{fig:dfg_example_intro}
\end{figure}

The shift from fine-granular PPM techniques, including next activity, remaining time, and outcome prediction, to model-based prediction, allows to obtain new insights into the global development of the process.
Consider the example in Figure~\ref{fig:dfg_example_intro} where the \href{https://doi.org/10.4121/uuid:270fd440-1057-4fb9-89a9-b699b47990f5}{road fine traffic management event log} is partitioned into 100 intervals in which an equal number of DF relations occur.
The DFs in the first 50 intervals are used to predict the next 25 intervals.
The DFGs show how process model forecasting and change exploration can provide multiple unique insights at a glance:
\begin{enumerate}
    \item Compared to the initial 50 intervals the proportion of fines sent decreases in the later intervals;
    \item The proportion of penalties remains comparable between the first 50 and next 25 time intervals;
    \item The number of occurrences and arc weights between \emph{Create Fine} and \emph{Send Fine} are forecasted with reasonable error ($\pm$15\%);
    \item The arc weights of the ending activities are predicted with reasonable error ($\pm$15\%).
\end{enumerate}
\vspace{-5mm}
These results provide insight both in terms of the past and present model, see items (1)-(2), and the quality of forecasts between the actual and forecasted model, (3)-(4).
Being able to construct such forecasts allows stakeholders to make estimates regarding how the overall fine system will evolve and allows to answer questions such as ``How many more fines will be received?'', ``Will the backlog of fines be reduced?'', ``Will all fines be paid'', and `Will the ratio of unpaid fines stay the same?'
This motivating example shows that, where process mining focuses on learning the as-is model to reason about trajectories of future cases and suggest potential repairs and improvements, process model forecasting allows to grasp the future stage of the overall process model in terms of a will-be model. % outcomes of the current as-is process, which allows to shortcut potentially wrong outcomes. 
%\todo[inline]{I found this shortcut-wrong statement hand wavy and misleading. I deleted it.}

A suitable means to evaluate the forecasts quantitatively is entropic relevance \cite{DBLP:conf/icpm/PolyvyanyyMG20}. This measure captures the quality of the discovered and forecasted DFGs with respect to the event logs they represent. 
Entropic relevance penalises the discrepancies in the relative frequencies of traces recorded in the log and described by the DFG as it stands for the average number of bits used to encode a log trace using the DFG, with small values being preferable to large ones.
If the entropic relevance of the forecasted DFG and the actual future DFG with respect to the test log is the same, then both DFGs represent the future behaviour similarly well. 
The entropic relevance of the historical DFG derived from the training log with respect to the testing log is 6.66 as indicated in Figure \ref{fig:dfg_example_intro}, 
%\todo[inline]{you refer to Fig 1 here, right? Please make that explicit.}
suggesting that the future behaviour shifts and the historical DFG still represents the behaviour in the log better than both the actual and forecasted DFGs which sit at an entropic relevance of 11.11. 

Measurement values are not enough to fully reveal the change of behaviour to the analyst. To this end, we complement the model-level prediction technique with a visualisation system to enable analysts to understand the forthcoming changes to the processes. Various process analysis tasks benefit from process forecasting~\cite{DBLP:conf/bpm/PollPRRR18}; most notably process forecasting helps understanding the incremental changes and adaptations that happen to the process model and to project them into the future. In terms of visualisation principles, we follow the ``Visual Information-Seeking Mantra":~\emph{overview first, zoom and filter, then details-on-demand}~\cite{DBLP:conf/vl/Shneiderman96}. 
%(maybe talk about tasks? not requirements)
Thus, we expect the design of our visualisation system to assist in the following tasks:

\begin{requidescr}
	\item[Identify process adaptations:\namedlabel{req:adaptation}] The visualisation system should assist the user in identifying the changes that happen in the process model of the future in respect to the past;
	\item[Allow for interactive exploration:\namedlabel{req:interactive}] The user should be able to follow the visual information-seeking principles, including overview first, filtering, zooming, and details-on-demand.
\end{requidescr} % CUSTOM from CDC, with love :)

Forecasting entire process models provides a new perspective on predictive process monitoring. 
%Observe that our process model forecasting technique intrinsically puts forwards an entirely new perspective on predictive process monitoring. 
The forecast horizon is substantially longer as compared to what existing next-activity prediction models can achieve. Moreover, where next activity and related PPM techniques have a strong case-level focus, a forecast at the model level provides a more comprehensive picture of the future development of the process.

%% file: 3-methodology.tex
\section{Process model forecasting}\label{sec:methodology}
This section outlines how time series of directly-follows relationships are extracted from event logs as well as how they are used to obtain process model forecasts with a range of widely-used forecasting techniques.
Finally, the visualisation of such forecasts is introduced.

\input{3.1-2-dftimeseries}

\input{3.3-visualisation}

%% file: 3.1-2-dftimeseries.tex
\subsection{From event log to directly-follows time series}\label{sec:3a:preliminaries}

An event log $L$ contains the recording of traces $\sigma \in L$ which are sequences of events produced by an information system during its execution.
A trace $\sigma=\langle e_1,...,e_{|\sigma|}\rangle \in \Sigma^*$ is a finite sequence over the alphabet of activities $\Sigma$ which serves as the set of event types.
Directly-follows relations between activities in an event log can be expressed as counting functions over activity pairs $>_L: \Sigma\times\Sigma \to \mathbb{N}$ so $>_L(a_1,a_2)$ counts the number of times activity $a_1$ is immediately followed by activity $a_2$ in the event log $L$.
Directly-follows relations can be calculated over all traces or a subset of subtraces of the log.
Finally, a Directly-Follows Graph (DFG) of the process then is the weighted directed graph with the activities as nodes and DF relations as weighted edges, i.e., $DFG=(\Sigma,>_L)$.

In order to obtain forecasts regarding the evolution of the DFG we construct DFGs for subsets of the log.
Many aggregations and bucketing techniques exist for next-step, performance, and goal-oriented outcome prediction \cite{DBLP:conf/caise/TaxVRD17,DBLP:journals/tkdd/TeinemaaDRM19,nguyen2016business}, e.g., predictions at a point in the process rely on prefixes of a certain length, or particular state aggregations \cite{DBLP:journals/sosym/AalstRVDKG10}.
In the forecasting approach proposed here, we integrate concepts from time-series analysis.
Hence, the evolution of the DFGs is monitored over intervals of the log where multiple aggregations are possible:
\begin{itemize}
	\item \textbf{Equitemporal aggregation:} each sublog $L_s\in L$  of interval $s$ contains a part of the event log of some fixed time duration. This can lead to sparsely populated sublogs when the events' occurrences are not uniformly spread over time; however, it is easy to apply on new traces.
	\item \textbf{Equisized aggregation:} each sublog $L_s\in L$ of interval $s$ contains a part of the event log where an equal amount of DF pairs occurred which leads to well-populated sublogs when enough events are available.
\end{itemize}
Tables \ref{tab:eventlog} and \ref{tab:aggregation} exemplify the aggregations.
These aggregations are useful for the following reasons.
First, an equisized aggregation, in general, has a higher likelihood of the underlying DFs approaching a white noise time series which is required for a wide range of time series forecasting techniques \cite{hyndman2018forecasting}. 
Second, both offer different thresholds at which forecasting can be applied.
In the case of the equisized aggregation, it is easier to quickly construct a desired number of intervals by simply dividing an event log into the equisized intervals.
However, most time series forecasting techniques rely on the time intervals being of equal duration which is embodied into the equitemporal aggregation \cite{kil1997optimum}.
Time series for the DFs $>_{T_{a_1,a_2}}=\langle >_{L_1}(a_1,a_2),\dots,>_{L_s}(a_1,a_2)\rangle, \forall a_1,a_2\in \Sigma\times\Sigma$ can be obtained for all activity pairs where $\bigcup^{L_s}_{L_1}=L$ by applying the aforementioned aggregations to obtain the sublogs for the intervals.
\begin{table}[htbp]
\centering
   \begin{minipage}{.45\textwidth}
   	\centering
   	% \resizebox{0.6\textwidth}{!}{
    \begin{tabular}{|l|l|l|}
    \toprule
    {Case ID} & Activity &Timestamp \\
    \midrule
    1     & $a_1$    & 11:30 \\
    1     & $a_2$    & 11:45 \\
    1     & $a_1$    & 12:10 \\
    1     & $a_2$    & 12:15 \\
    \midrule
    2     & $a_1$    & 11:40 \\
    2     & $a_1$    & 11:55 \\
    \midrule
    3     & $a_1$    & 12:20 \\
    3     & $a_2$    & 12:40 \\
    3     & $a_2$    & 12:45 \\
    \bottomrule
    \end{tabular}
    % }
    \caption{Example event log with 3 traces and 2 activities.}
\label{tab:eventlog}
\end{minipage}\hfill
  \begin{minipage}{.45\textwidth}
  	\centering
%   	\resizebox{0.75\textwidth}{!}{
    \begin{tabular}{|l|c|c|}
    \toprule
    DF    & Equitemporal  & Equisized \\
    \midrule
    $<_{Ls}(a_1,a_1)$ & (0,1,0) & (1,0,0) \\
    $<_{Ls}(a_1,a_2)$ & (1,1,1) & (1,1,1) \\
    $<_{Ls}(a_2,a_1)$ & (0,1,0) & (0,1,0) \\
    $<_{Ls}(a_2,a_2)$ & (0,0,1) & (0,0,1) \\
    \bottomrule
    \end{tabular}
    % }
  \caption{An example of using an interval of 3 used for equitemporal aggregation (75 minutes in 3 intervals of 25 minutes) and equisized intervals of size 2 (6 DFs over 3 intervals)).}
  \label{tab:aggregation}
 \end{minipage}%
\end{table}%
\vspace{-15mm}
\subsection{From DF time series to process model forecasts}\label{sec:3b:df}

The goal of process model forecasting is to obtain a forecast for future DFGs by combining the forecasts of all the DF time series.
To this purpose, we propose to use time series techniques to forecast the DFG at time $T+h$ given time series up until $T$ $\widehat{DFG}_{T+h}=(\Sigma,\{\hat{>}_{T+h|T_{a_1,a_2}}|a_1,a_2\in \Sigma\times\Sigma\})$ for which various algorithms can be used.
In time series modelling, the main objective is to obtain a forecast $\hat{y}_{T+h|T}$ for a horizon $h\in \mathbb{N}$ based on previous $T$ values in the series $(y_1,...,y_T)$ \cite{hyndman2018forecasting}.
For example, the naive forecast simply uses the last value of the time series $T$ as its forecast $\hat{y}_{T+h|T}=y_T$.
An alternative naive forecast uses the average value of the time series $T$ as its forecast $\hat{y}_{T+h|T}=\frac{1}{T}\Sigma_i^{T} y_i$.

A trade-off exists between approaching DFGs as a multivariate collection of DF time series, or treating each DF separately.
Traditional time series techniques use univariate data in contrast with multivariate approaches such as Vector AutoRegression (VAR) models, and machine learning-based methods such as neural networks or random forest regressors.
Despite their simple setup, it is debated whether machine learning methods necessarily outperform traditional statistical approaches.
The study in \cite{makridakis2018statistical} found that this is not the case on a large number of datasets and the authors note that machine learning algorithms require significantly more computational power.
This result was later reaffirmed, although it is noted that hybrid solutions are effective \cite{makridakis2020m4}.
For longer horizons, traditional time series approaches still outperform machine learning-based models.
Given the potentially high number of DF pairs in a DFG, the proposed approach uses a time series algorithm for each DF series separately.
VAR models would require a high number of intervals (at least as many as there are DFs times the lag coefficient) to estimate all parameters of all the time series despite their potentially strong performance \cite{thomakos2004naive}.
Machine learning models could potentially leverage interrelations between the different DFs but again would require training set way larger than typically available for process mining to account for dimensionality issues due to the potentially high number of DFs. 
Therefore, in this paper, traditional time series approaches are chosen and applied to the univariate DF time series, with at least one observation per sublog/time interval present.

Autoregressive, moving averages, ARIMA, and varying variance models make up the main families of traditional time series forecasting techniques\cite{hyndman2018forecasting}.
In addition, a wide array of other forecasting techniques exist, ranging from simple models such as naive forecasts over to more advanced approaches such as exponential smoothing and auto-regressive models.
Many also exist in a seasonal variant due to their application in contexts such as sales forecasting.

The Simple Exponential Smoothing (SES) model uses a weighted average of past values whose importance exponentially decays as they are further into the past, where the Holt's models introduce a trend in the forecast, meaning the forecast is not `flat'.
Exponential smoothing models often perform very well despite their simple setup \cite{makridakis2018statistical}.
AutoRegressive Integrating Moving Average (ARIMA) models are based on auto-correlations within time series. 
They combine auto-regressions with a moving average over error terms.
It is established by a combination of an AutoRegressive (AR) model of order $p$ to use the past $p$ values in the time series and to apply a regression over them and a Moving Average (MA) model of order $q$ to create a moving average of the past forecast errors.
Given the necessity of using a white noise series for AR and MA models, data is often differenced to obtain such series \cite{hyndman2018forecasting}.
ARIMA models then combine both AR and MA models where the integration occurs after modelling, as these models are fitted over differenced time series.
ARIMA models are considered to be one of the strongest time series modelling techniques \cite{hyndman2018forecasting}.
An extension to ARIMA, which is widely used in econometrics, are the (Generalized) AutoRegressive Conditional Heteroskedasticity ((G)ARCH) models \cite{francq2019garch}.
These models relax the assumption that the variance of the error term has to be constant over time, and rather model this variance as a function of the previous error term.
For AR-models, this leads to the use of ARCH-models, while for ARMA models GARCH-models are used as follows.
An ARCH(q) model captures the change in variance by allowing it to gradually increase over time or to allow for short bursts of increased variance.
A GARCH(p,q) model combines both the past values of observations and the past values of variance.
(G)ARCH models often outperform ARIMA models in contexts such as the forecast of financial indicators, in which the variance often changes over time \cite{francq2019garch}.

In general, we can regard linear SES models as a subset of ARIMA models, where (G)ARCH models are specializations of ARIMA models that can be regarded as increasingly complex and better capable of modelling particular intricacies in the time series.
However, the success of different models for forecasting purposes does not depend on their complexity, and the most suitable technique is mainly determined by performance on training and test sets.

%% file: 3.3-visualisation.tex
\subsection{Process change exploration}\label{sec:3c:pce}

% as has been shown the prediction is nice but it is difficult to derive the insignts 
% definition of the visualization
% humans cannot comprehend the results and we need visualzation 
% (new functions on vis that ar enot mentioned before should be mentioned?)

%%%
% outline:
% 0. short intro
% 1. user tasks
% 2. design (and what supports design)
% 3. implementation and the resources.
%%%

In Sections~\ref{sec:3a:preliminaries} and \ref{sec:3b:df} we described the approach for forecasting process models. To that end, gaining actual insights from such forecasted values remains a difficult task for the analyst. This section sets off to present the design of a novel visualisation system to aid analysts in the exploration of the event logs and their corresponding (forecasted) discovered process models.

Following the user tasks~\ref{req:adaptation} and~\ref{req:interactive} from~\Cref{sec:2:motivation}, we designed a Process Change Exploration (PCE) system to support the interpretation of the process model forecasts. PCE is an interactive visualisation system that consists of three connected views.

%In order to design the system we first established user tasks as a basis for the system design decisions. 

%To derive the user tasks we focus on the requirements of process mining analysis with respect to process forecasting and visualization principles. The authors of~\cite{DBLP:conf/bpm/PollPRRR18} discuss the opportunities for process forecasting. They describe that the utility of process forecasting is an understanding of the incremental changes or adaptations that happen to the process model into the future. In designing an explorative visualization system, we also followed the "Visual Information-Seeking Mantra:"~\emph{overview first, zoom and filter, then details-on-demand}~\cite{DBLP:conf/vl/Shneiderman96}. 
%%(maybe talk about tasks? not requirements)
%Thus, we expect the design of our system to assist in the following tasks:
%
%%requirements for the visualization based on the related literature and experience working with event sequence data. 
%
%\begin{requidescr}
%	\item[Identify process adaptations:\namedlabel{req:adaptation}] The visualization system should assist the user in identifying the changes that happen in the process model of the future in respect to the past;
%	\item[Allow for interactive exploration:\namedlabel{req:interactive}] The user should be able to follow the visual information-seeking principles, including overview first, filtering, zooming, details-on-demand principles.;
%\end{requidescr} % CUSTOM from CDC, with love :)

\textbf{Adaptation Directly-Follows Graph (aDFG) view.} This is the main view of the visualisation that will show the model of the process. In order to accomplish task~\ref{req:adaptation}, we modify the DFG syntax. To display the process model adaptation from time range $T_{i_0}-T_{j_0}, i_0<j_0$, to $T_{i_1}-T_{j_1}, i_1<j_1$, we display the union of the process models of these regions, annotating the nodes and edges with the numbers of both ranges. We colour the aDFG as follows: we use colour saturation to show the nodes with higher values. We colour edges with a diverging saturation (red-black-green) schema. This colouring applies red colour to edges that are dominant in the $T_{i_0}-T_{j_0}$ range, and green if edges are dominant in the $T_{i_1}-T_{j_1}$ range, otherwise the edge colour is close to black. For coloring edges, we reused the idea of the three colour schema from~\cite{DBLP:conf/grapp/KriglsteinR12}.

\textbf{Timeline view with brushed regions.} This view represents the area chart graph that shows how the number of activity executions changes with time. The colour of the the area chart is split into two parts, one for the actual data and the other to show the time range of forecasted values. 
Analysts can brush one region in order to zoom in, creating one region of interest $T_{i_0,}-T_{j_0}, i_0<j_0$ that is displayed on the DFG. Analysts can also brush two regions of the area chart to select two time ranges, updating the DFG to the aDFG representation. The brushed regions are coloured accordingly to the schema for colouring aDFG transitions. The earlier brushed region is coloured in red, while the second one is coloured green. 

\textbf{Activity and path sliders.} We adopt two sliders to simplify the DFG~\cite{leemans2019directly} and the aDFG for detailed exploration of the models.

Based on the described views, we conjecture that the analyst can accomplish tasks~\ref{req:adaptation} and~\ref{req:interactive} with ease.

%% file: 4-evaluation.tex
\section{Implementation and evaluation}\label{sec:experiment}
In this section, an experimental evaluation over six real-life event logs is reported.
The aim of the evaluation is to measure to what extent the forecasted DFG process models are capable of correctly reproducing actual future DFGs in terms of allowing for the same process model behaviour.
To this end, we benchmark the actual against the forecasted entropic relevance, as discussed in \Cref{sec:2:motivation}.
This is done for various parts of the log, i.e. forecasts for the middle time spans of the event logs up to the later parts of the event log to capture the robustness of the forecasting techniques in terms of the amount of data required to obtain good results for both the equisized and equitemporal aggregation.

\subsection{Re-sampling and test setup}
To obtain training data, time series are constructed by specifying the number of intervals (i.e., time steps in the DF time series) using either equitemporal or equisized aggregation, as described in Section \ref{sec:3a:preliminaries}.
Time series algorithms are parametric and sensitive to sample size requirements \cite{hanke2001business}.
Depending on the number of parameters a model uses, a minimum size of at least 50 steps is not uncommon.
However, typically, model performance should be monitored at a varying number of steps.
In the experimental evaluation, the event logs are divided into 100 time intervals with a varying share of training and test intervals. A constant and long horizon $h=25$ is used meaning all test sets contain 25 intervals, but the training sets are varied from $ts=25$ to $ts=75$ intervals; the forecasts progressively target the forecast of intervals 25-50 (the second quarter of intervals) over to 75-100 (the last quarter of intervals).
This allows us to inspect the difference in results when only a few data points are used, or data points in the middle or towards the end of the available event data are used.

Resampling is applied based on 10-fold cross-validation constructed following a rolling window approach for all horizon values $h\in[1,25]$ where a recursive strategy is used to iteratively obtain $\hat{y}_{t+h|T_{t+h-1}}$ with $(y_1,\dots,y_{T},\dots,\hat{y}_{t+h-1})$ \cite{weigend2018time}.
Ten training sets are hence constructed for each training set length $ts$ and range from $(y_1,\dots,y_{T-h-f})$ and the test sets from $(y_{T-h-f+1},\dots,y_{T-f})$ with $f\in[0,9]$ the fold index \cite{bergmeir2012use}.
While direct strategies with a separate model for every value of $h$ can be used as well and avoid the accumulation of error, they do not take into account statistical dependencies for subsequent forecasts.

Six often-used, publicly available event logs are used: the \href{https://doi.org/10.4121/uuid:3926db30-f712-4394-aebc-75976070e91f}{BPI challenge of 2012 log}, \href{https://doi.org/10.4121/uuid:5f3067df-f10b-45da-b98b-86ae4c7a310b}{2017}, and \href{https://doi.org/10.4121/uuid:3301445f-95e8-4ff0-98a4-901f1f204972}{2018}, the \href{https://doi.org/10.4121/uuid:915d2bfb-7e84-49ad-a286-dc35f063a460}{Sepsis cases event}, \href{https://doi.org/10.4121/uuid:0c60edf1-6f83-4e75-9367-4c63b3e9d5bb}{an Italian help desk}, and the Road Traffic Fine Management Process log (RTFMP) event log.
Each of these logs has a diverse set of characteristics in terms of case and activity volume and average trace length, as shown in Table \ref{tab:eventlogs}.

\begin{table}[htbp]
  \centering
  \resizebox{0.6\textwidth}{!}{
    \begin{tabular}{lrrr}
    \toprule
    \textbf{Event log} & \multicolumn{1}{l}{\textbf{\# cases}} & \multicolumn{1}{l}{\textbf{\# activities}} & \multicolumn{1}{l}{\textbf{Average trace length}} \\
    \midrule
    \textbf{BPI 12} & 13,087 & 36    & 20.02 \\
    \textbf{BPI 17} & 31,509 & 26    & 36.83 \\
    \textbf{BPI 18} & 43,809 & 170   & 57.39 \\
    \textbf{Sepsis} & 1,050 & 16    & 14.49 \\
    \textbf{RTFMP} & 150,370 & 11    & 3.73 \\
    \textbf{Italian} & 4,580 & 14    & 4.66 \\
    \bottomrule
    \end{tabular}%
    }
  \caption{Overview of the characteristics of the event logs used in the evaluation.}
  \label{tab:eventlogs}%
\end{table}%

%An example of applying the equisized or equitemporal aggregation to the Sepsis event log with 100 intervals results in the DF time series of Figure \ref{fig:sepsists}, where the DF occurrences of the most frequently occurring activity pair is included.
%For the equisized aggregation, the number of DFs is indeed relatively stable over the log's timeline where for the equitemporal aggregation a noticeable decline of DF pairs is visible towards the end of the series.
%This phenomenon is typical in event logs, as processes usually have particular endpoint activities, but can also be due to the unequal distribution of events over the event log's time line.
\vspace{-1cm}
There are a few considerations concerning the DF time series in these event logs.
Firstly, DFs of activity pairs containing endpoint activities (i.e. at the start/end of a trace) often only contain meaningful numbers at very particular parts of the series and are hard to process by longitudinal algorithms which require a more extended pattern to extract a meaningful pattern for foresting.
Secondly, the equitemporal aggregation can suffer from event logs in which events do not occur frequently throughout the complete log's timespan.
For instance, the Sepsis log's number of event occurrences tails off towards the end which can be alleviated by pre-processing (not done here to remain consistent over the event logs).
Finally, suppose the level of occurrences of the DF pairs is low and close to zero. In that case, the series might be too unsuitable for analysis using white noise series analysis techniques that assume stationarity.
Ideally, every time series should be evaluated using a stationarity test such as the Dickey-Fuller unit root test \cite{leybourne1995testing}, and an appropriate lag order established for differencing to ensure a white noise process is used for training. 
Furthermore, for each algorithm, especially for ARIMA-based models, (partial) auto-correlation has to be established to obtain the ideal $p$ and $q$ parameters.
However, for the sake of simplicity and to avoid solutions where each activity pair has to have different parameters, various values are used for $p$, $d$, and $q$ and applied to all DF pairs where only the best-performing are reported below for comparison with the other time series techniques.
The results contain the best-performing representative of each forecasting family.
% \begin{figure}[tb]
% 	\centering
% 	\subfigure[Most common DF - equisize]{\includegraphics[width=0.39\textwidth]{./img/rtfmp_1.png}}
% 	\subfigure[Most common DF - equitemp]{\includegraphics[width=0.39\textwidth]{./img/rtfmp_1_t.png}}
% 	\caption{RTFMP}
% 	\label{fig:sepsists}
% \end{figure}

% \subsection{Evaluation criteria}
% Given that we want to evaluate the capability of the approach to accurately predict the evolution of the process model, the combination of all DF predictions is considered to obtain a global DFG prediction.
% The following two criteria are used:
% \begin{itemize}
% 	\item \textbf{Cosine distance:} measures the distance between two vectors and is often used to compare graph distance. This metric is used to compare the DFGs' edge weight matrices between the actual and predicted number of DF relations.
% 	\item \textbf{Entropic relevance:} see section \ref{sec:2:motivation}.
% \end{itemize}
% These criteria balance a predictive and structural evaluation of the algorithms and report on both the numeric performance common in a forecasting setting as well as their appropriateness in terms of reproducing a structurally usable process model which allows for the observed process behaviour.
% In both cases a lower score is better.

\subsection{Results}\label{sec:4.3:results}
All pre-processing was done in Python with a combination of \emph{pm4py}\footnote{\url{https://pm4py.fit.fraunhofer.de}} and the \emph{statsmodels} package \cite{seabold2010statsmodels}. 
The code is publicly available.\footnote{\url{https://github.com/JohannesDeSmedt/pmf}}

To get a grasp of the forecasting performance in combination with the actual use of DFGs (which are rarely used in their non-aggregated form \cite{van2019practitioner}) we present the mean absolute percentage error (MAPE) between the entropic relevance of the actual and forecasted DFGs at both full size, at 50\%, and 75\% reduction which is node-based (i.e. only the Q2/Q3 percentile of nodes in terms of frequency is retained).
Hence, we obtain a measure of accuracy in terms of the discrepancy of the actual and forecasted model behaviour.
Using different levels of aggregation also balances recall and precision, as aggregated DFGs are less precise but possibly less overfitting.
The results can be found in Tables \ref{tab:result_dfg_table} to \ref{tab:result_dfg_table_25}.
NAs are reported when the algorithms did not converge, no data was available (e.g. Sepsis for the 75-100 equitemporal intervals), or extremely high values were forecasted.

When no reduction is applied, Table \ref{tab:result_dfg_table} shows that for the BPI12 and BPI17 logs, a below 10\% error can be achieved, primarily for equisized aggregation. 
For the Italian help desk log, results are in the 10-37\% bracket, while for the other logs, results are often well above a 100\% deviation (with the entropic relevance of the actual DFGs being lower, hence better, than the entropic relevance of the forecasted DFGs).
However, for the RTFMP and BPI18 log, results are better when more training points are used (e.g., 50 or 75 to obtain forecasts for the 50-75 and 75-100 intervals).
There is no significant difference between equisized and equitemporal aggregation except for the occasional outliers.
Overall, the percentage error is lower in Table \ref{tab:result_dfg_table_50} when a reduction of 50\% is applied with sub-10\% results for the BPI12, Sepsis, and BPI17 logs. 
The results for the RTFMP log are occasionally better but mostly worse, similar to BPI18.
Finally, the results in Table \ref{tab:result_dfg_table_25} show a further reduction of errors for the BPI12, Sepsis, BPI17, and Italian logs and a drastic decrease to close to 0\% for RTFMP.
The results for the BPI18 log remain bad at over 100\% error rates.

These results are commensurate with the findings in \cite{DBLP:conf/icpm/PolyvyanyyMG20}, which contains entropic relevance results for the BPI12, Sepsis, and RTFMP logs, indicating that entropic relevance of larger DFGs is lower (better) for RTFMP/Sespsis, and the entropic relevance goes up strongly for small models of RTFMP meaning the drastically improved error rates reported here are for models performing worse in terms of recall and precision.
The entropic relevance for the BPI12 log is stable for the full spectrum of DFG sizes as per \cite{DBLP:conf/icpm/PolyvyanyyMG20}, which is reflected in the consistently good error rates presented here.
This means that the low error rates reported are produced by the reduced DFGs, which still score strongly in terms of recall and precision.
Matching all results to the event log characteristics, we notice that the event logs with longer traces with medium-sized alphabets ($>$20) such as BPI12 and BPI17 consistently report good results.
The BPI18 log's high number of activities seems to inflate error rates quickly, which is further aggravated when DFGs are reduced.
Given that DFGs are based on activity pairs, this result is not surprising.
For Sepsis and the Italian event logs, good error rates are obtained once DFGs are reduced, indicating that forecasting the low-frequent edges and activities might lead to high error rates when the alphabet is smaller and traces are shorter, which is potentially also caused by the lack of precision as witnessed with the RTFMP log.

Overall, there exist many scenarios in which process model forecasting is delivering solid results.
For the BPI12, BPI17, Italian, and Sepsis event logs, sub-10\% error rates can be achieved both for equisized and equitemporal aggregation combined with model reductions which readers of DFGs typically apply.
In some cases, even a naive forecast is enough to obtain a low error rate.
However, the AR and ARIMA models report the best error rates in most cases.
Nevertheless, results are often close except when fewer training points are used.
Then, results are often varying widely.
In future work, the robustness of the forecast algorithms will be further investigated, e.g., via scrutinising the confidence intervals of the forecasted DF outcomes.
\vspace{-5mm}
\input{tables/table_dfg.tex}
\vspace{-17mm}
\input{tables/table_dfg_50.tex}
\input{tables/table_dfg_25.tex}

\input{4.5-visualisation}

%% file: tables/table_dfg.tex
\begin{table}[ht]
  \centering
  \resizebox{\textwidth}{!}{
    \begin{tabular}{|c|l|ccc|ccc|ccc|ccc|ccc|ccc|}
\cmidrule{3-20}    \multicolumn{1}{r}{} &       & \multicolumn{3}{c|}{\textbf{BPI 12}} & \multicolumn{3}{c|}{\textbf{Sepsis}} & \multicolumn{3}{c|}{\textbf{RTFMP}} & \multicolumn{3}{c|}{\textbf{BPI17}} & \multicolumn{3}{c|}{\textbf{Italian}} & \multicolumn{3}{c|}{\textbf{BPI18}} \\
    \multicolumn{1}{r}{} &       & \textbf{50} & \textbf{75} & \textbf{100} & \textbf{50} & \textbf{75} & \textbf{100} & \textbf{50} & \textbf{75} & \textbf{100} & \textbf{50} & \textbf{75} & \textbf{100} & \textbf{50} & \textbf{75} & \textbf{100} & \textbf{50} & \textbf{75} & \textbf{100} \\
    \midrule
    \multirow{5}[2]{*}{\begin{sideways}\textbf{equisize}\end{sideways}} & \textbf{nav} & 9.74  & 8.56  & 9.82  & 97.09 & 97.40 & 100.76 & 437.14 & \textbf{105.81} & 115.34 & 6.86  & 8.80  & \textbf{7.00} & 25.93 & 16.52 & 37.71 & 82.10 & \textbf{99.90} & 38.41 \\
          & \textbf{arima212} & 12.41 & 9.75  & 10.80 & NA    & \textbf{83.31} & \textbf{100.58} & \textbf{398.66} & NA    & NA    & 10.03 & \textbf{8.54} & 13.23 & 24.60 & 9.17  & 39.01 & 82.81 & NA    & 30.12 \\
          & \textbf{ar2} & NA    & \textbf{8.45} & \textbf{9.62} & \textbf{97.04} & 97.40 & 100.76 & NA    & NA    & \textbf{110.14} & 6.83  & 14.84 & 13.83 & 23.81 & 13.98 & \textbf{36.89} & \textbf{78.82} & NA    & NA \\
          & \textbf{hw} & \textbf{8.61} & 8.96  & 10.14 & 97.09 & 97.40 & 100.76 & 402.83 & 110.17 & 130.10 & \textbf{6.81} & 8.68  & 186.94 & \textbf{22.54} & \textbf{9.14} & 43.31 & 81.04 & NA    & NA \\
          & \textbf{garch} & 11.47 & 8.60  & 10.17 & 97.09 & 97.40 & 100.76 & 426.71 & 109.79 & 117.15 & 6.89  & 8.82  & 186.94 & 25.48 & 31.29 & 65.54 & 72.89 & NA    & \textbf{28.59} \\
    \midrule
    \multicolumn{1}{|c|}{\multirow{5}[2]{*}{\begin{sideways}\textbf{equitemp}\end{sideways}}} & \textbf{nav} & 15.57 & 10.14 & 12.63 & 98.51 & 100.75 & NA    & 199.69 & 29.70 & 36.15 & 7.12  & 8.63  & \textbf{13.41} & 27.12 & 26.86 & 39.94 & NA    & NA    & 54.57 \\
          & \textbf{arima212} & NA    & 11.67 & \textbf{12.00} & \textbf{89.07} & \textbf{100.39} & NA    & \textbf{122.63} & \textbf{28.55} & \textbf{33.82} & 8.13  & 158.70 & 18.74 & 26.59 & 24.26 & 38.03 & NA    & \textbf{42.83} & NA \\
          & \textbf{ar2} & NA    & \textbf{9.97} & 12.43 & 98.37 & 100.75 & NA    & NA    & 29.74 & NA    & 7.09  & NA    & 19.60 & \textbf{26.33} & 30.02 & 38.68 & NA    & NA    & NA \\
          & \textbf{hw} & \textbf{13.09} & 10.46 & 12.08 & 98.40 & 100.75 & NA    & 162.94 & 29.34 & 36.15 & \textbf{7.07} & \textbf{8.35} & 186.91 & 26.90 & \textbf{23.57} & \textbf{36.20} & NA    & 43.02 & NA \\
          & \textbf{garch} & 17.80 & 10.29 & 12.71 & 95.75 & 100.75 & NA    & 199.13 & 30.44 & 36.00 & 7.37  & 187.45 & 186.91 & 27.11 & 45.58 & 55.67 & NA    & 46.69 & \textbf{42.97} \\
    \bottomrule
    \end{tabular}%

    }
    \caption{Overview of the mean percentage error in terms of entropic relevance for the full DFGs.}
  \label{tab:result_dfg_table}%
\end{table}%

%% file: tables/table_dfg_50.tex
\begin{table}[ht]
  \centering
  \resizebox{\textwidth}{!}{
    \begin{tabular}{|c|l|ccc|ccc|ccc|ccc|ccc|ccc|}
\cmidrule{3-20}    \multicolumn{1}{r}{} &       & \multicolumn{3}{c|}{\textbf{BPI 12}} & \multicolumn{3}{c|}{\textbf{Sepsis}} & \multicolumn{3}{c|}{\textbf{RTFMP}} & \multicolumn{3}{c|}{\textbf{BPI17}} & \multicolumn{3}{c|}{\textbf{Italian}} & \multicolumn{3}{c|}{\textbf{BPI18}} \\
    \multicolumn{1}{r}{} &       & \textbf{50} & \textbf{75} & \textbf{100} & \textbf{50} & \textbf{75} & \textbf{100} & \textbf{50} & \textbf{75} & \textbf{100} & \textbf{50} & \textbf{75} & \textbf{100} & \textbf{50} & \textbf{75} & \textbf{100} & \textbf{50} & \textbf{75} & \textbf{100} \\
    \midrule
    \multirow{5}[2]{*}{\begin{sideways}\textbf{equisize}\end{sideways}} & \textbf{nav} & \textbf{4.65} & \textbf{5.83} & \textbf{11.50} & \textbf{8.35} & 8.80  & 6.29  & 234.18 & 295.99 & 203.68 & 7.82  & 9.22  & 11.04 & 23.05 & 14.18 & 21.66 & \textbf{252.76} & \textbf{231.44} & \textbf{160.66} \\
          & \textbf{arima212} & 7.96  & 22.89 & 13.43 & 8.55  & 8.81  & \textbf{6.14} & 234.14 & 288.27 & 198.86 & 4.49  & 5.98  & 10.67 & 22.31 & 7.32  & 23.17 & 369.47 & 252.24 & 218.93 \\
          & \textbf{ar2} & 24.53 & 27.58 & 30.81 & 8.54  & 8.72  & 6.30  & 234.57 & 293.10 & 201.21 & \textbf{4.27} & 6.08  & \textbf{10.22} & 21.26 & 11.91 & \textbf{20.56} & NA    & 230.02 & NA \\
          & \textbf{hw} & 45.80 & 38.02 & 13.13 & 8.73  & \textbf{8.65} & 6.17  & 233.05 & \textbf{151.19} & \textbf{111.89} & 4.51  & \textbf{5.37} & 11.06 & \textbf{20.33} & \textbf{7.28} & 26.12 & 391.38 & 280.59 & 226.27 \\
          & \textbf{garch} & 26.30 & 23.77 & 48.86 & 8.63  & 8.87  & 7.06  & \textbf{231.70} & 295.93 & 203.45 & 4.50  & 9.41  & 11.09 & 23.18 & 29.07 & 45.31 & 315.99 & 234.79 & 217.62 \\
    \midrule
    \multicolumn{1}{|c|}{\multirow{5}[2]{*}{\begin{sideways}\textbf{equitemp}\end{sideways}}} & \textbf{nav} & \textbf{7.15} & \textbf{6.86} & 17.86 & 6.41  & \textbf{7.73} & NA    & \textbf{75.48} & \textbf{36.18} & \textbf{86.46} & 5.93  & 7.23  & 30.98 & 24.35 & 18.64 & 26.48 & NA    & \textbf{219.13} & 410.13 \\
          & \textbf{arima212} & 49.87 & 10.59 & 19.59 & \textbf{4.91} & 8.13  & NA    & 135.97 & 40.22 & 86.74 & \textbf{5.64} & \textbf{5.13} & \textbf{30.30} & \textbf{23.48} & 16.32 & \textbf{20.45} & \textbf{205.21} & 261.81 & \textbf{253.38} \\
          & \textbf{ar2} & 21.06 & 7.49  & 18.85 & 7.17  & 8.08  & NA    & 95.44 & 36.30 & 86.60 & 5.70  & NA    & 30.97 & 23.67 & 21.71 & 25.44 & NA    & NA    & 443.76 \\
          & \textbf{hw} & 7.41  & 7.02  & \textbf{17.54} & 6.72  & 10.34 & NA    & 77.93 & 36.62 & 86.76 & 5.95  & 7.39  & 30.86 & 23.55 & \textbf{15.66} & 22.91 & NA    & 236.37 & 439.04 \\
          & \textbf{garch} & 57.44 & 32.85 & 37.98 & 6.76  & 7.85  & NA    & \textbf{75.48} & 36.20 & 86.52 & 5.93  & 7.33  & 31.12 & 24.24 & 36.51 & 40.40 & NA    & 283.40 & 492.85 \\
    \bottomrule
    \end{tabular}%
    }
    \caption{Overview of the mean percentage error in terms of entropic relevance for the DFGs with a 50\% reduction.}
  \label{tab:result_dfg_table_50}%
\end{table}%

%% file: tables/table_dfg_25.tex
\begin{table}[ht]
  \centering
  \resizebox{\textwidth}{!}{
    \begin{tabular}{|c|l|ccc|ccc|ccc|ccc|ccc|ccc|}
\cmidrule{3-20}    \multicolumn{1}{r}{} &       & \multicolumn{3}{c|}{\textbf{BPI 12}} & \multicolumn{3}{c|}{\textbf{Sepsis}} & \multicolumn{3}{c|}{\textbf{RTFMP}} & \multicolumn{3}{c|}{\textbf{BPI17}} & \multicolumn{3}{c|}{\textbf{Italian}} & \multicolumn{3}{c|}{\textbf{BPI18}} \\
    \multicolumn{1}{r}{} &       & \textbf{50} & \textbf{75} & \textbf{100} & \textbf{50} & \textbf{75} & \textbf{100} & \textbf{50} & \textbf{75} & \textbf{100} & \textbf{50} & \textbf{75} & \textbf{100} & \textbf{50} & \textbf{75} & \textbf{100} & \textbf{50} & \textbf{75} & \textbf{100} \\
    \midrule
    \multirow{5}[2]{*}{\begin{sideways}\textbf{equisize}\end{sideways}} & \textbf{nav} & 0.96  & 0.87  & 1.37  & 2.40  & 2.91  & \textbf{3.47} & 0.00  & 0.00  & 0.01  & 0.12  & \textbf{0.29} & \textbf{0.14} & 13.11 & 15.11 & 27.02 & \textbf{247.32} & 223.34 & \textbf{146.27} \\
          & \textbf{arima212} & 0.98  & \textbf{0.84} & \textbf{1.33} & \textbf{2.38} & 2.85  & 3.65  & 0.00  & 0.00  & 0.01  & \textbf{0.06} & 0.30  & 0.16  & 13.08 & 14.94 & 26.97 & 346.94 & 236.73 & 211.17 \\
          & \textbf{ar2} & \textbf{0.86} & 0.85  & 1.35  & 2.42  & 2.58  & \textbf{3.47} & 0.00  & 0.00  & 0.01  & 0.12  & 0.30  & \textbf{0.14} & 13.10 & 15.02 & 26.97 & NA    & \textbf{222.06} & NA \\
          & \textbf{hw} & 1.00  & 0.85  & 1.35  & 2.73  & 2.90  & \textbf{3.47} & 0.00  & 0.00  & 0.01  & 0.11  & 0.31  & 0.15  & \textbf{12.98} & \textbf{14.79} & 27.08 & 333.80 & 255.83 & 207.86 \\
          & \textbf{garch} & 0.89  & 0.86  & 1.35  & 2.51  & \textbf{2.49} & \textbf{3.47} & 0.00  & 0.00  & 0.01  & 0.12  & \textbf{0.29} & \textbf{0.14} & 13.13 & 15.00 & \textbf{26.89} & 299.59 & 248.75 & 182.07 \\
    \midrule
    \multicolumn{1}{|c|}{\multirow{5}[2]{*}{\begin{sideways}\textbf{equitemp}\end{sideways}}} & \textbf{nav} & 4.92  & 3.55  & 4.05  & \textbf{2.31} & \textbf{1.65} & NA    & 0.03  & 0.02  & 0.11  & \textbf{0.05} & \textbf{0.11} & 5.92  & 8.18  & 30.10 & 20.61 & NA    & \textbf{203.92} & 562.09 \\
          & \textbf{arima212} & 4.93  & 3.59  & 3.86  & 2.77  & 2.62  & NA    & 0.03  & 0.02  & 0.11  & 0.11  & 0.14  & \textbf{5.82} & 8.39  & 30.20 & 20.54 & \textbf{180.36} & 245.18 & \textbf{191.25} \\
          & \textbf{ar2} & 4.85  & \textbf{3.52} & 4.04  & 2.35  & 2.93  & NA    & 0.03  & 0.02  & 0.11  & 0.06  & NA    & 5.91  & 8.19  & 30.17 & 20.58 & NA    & NA    & 559.14 \\
          & \textbf{hw} & \textbf{4.82} & \textbf{3.52} & 3.84  & 2.47  & 1.67  & NA    & 0.03  & 0.02  & 0.11  & 0.06  & 0.13  & 5.85  & 8.45  & \textbf{20.00} & \textbf{15.90} & NA    & 228.78 & 384.37 \\
          & \textbf{garch} & 7.97  & 3.54  & 4.02  & 2.34  & 2.92  & NA    & 0.03  & 0.02  & 0.11  & \textbf{0.05} & 0.12  & 5.93  & \textbf{8.17} & 30.03 & 20.52 & NA    & 226.58 & 606.03 \\
    \bottomrule
    \end{tabular}%
    }
    \caption{Overview of the mean percentage error in terms of entropic relevance for the DFGs with a 75\% reduction.}
  \label{tab:result_dfg_table_25}%
\end{table}%

%% file: 4.5-visualisation.tex
\vspace{-1.5cm}
\subsection{Visualising Process Model Forecasts}\label{sec:visualisation}

In~\Cref{sec:4.3:results}, we evaluated forecasting results, ensuring the conformance and interpretability of the predicted process models. To that end, gaining insights from such predicted data remains a difficult task for the analyst. 
This section sets off to present a novel visualisation system to aid analysts in exploring the event logs. The process of designing and implementing the system started by designing several prototypes that undergone rounds of discussions to mature into the implemented visualisation system. 

\begin{figure}
	\centering
	\includegraphics[width=\textwidth]{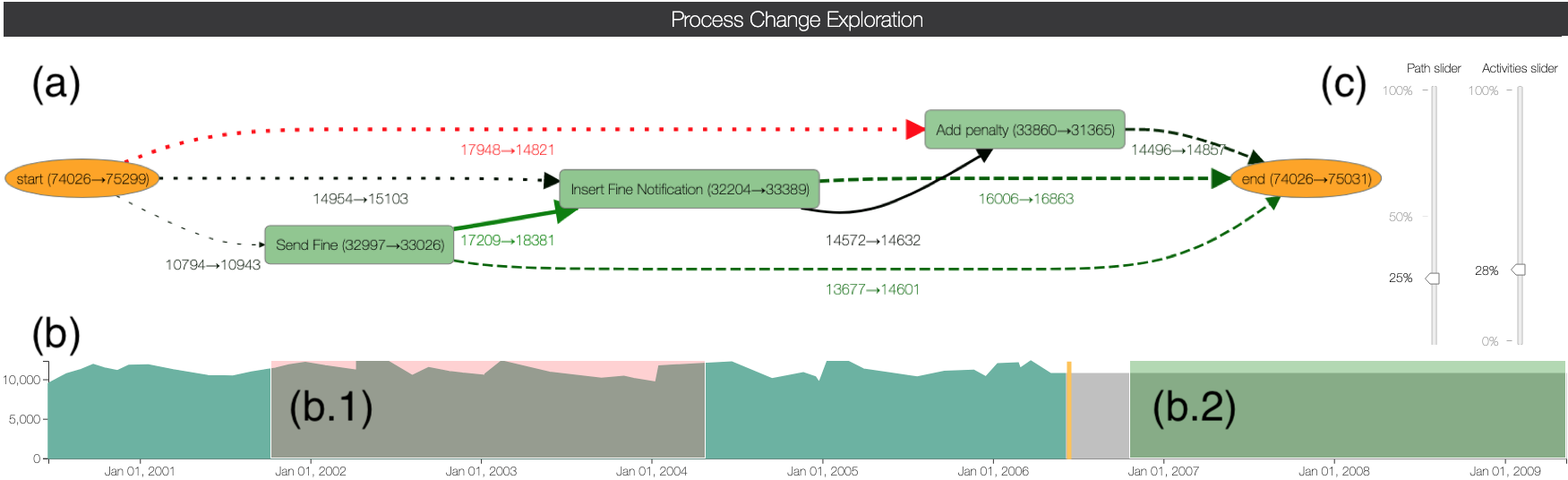}
	\caption{Process Change Exploration (PCE) system.~\emph{(a)} shows~\emph{Adaptation Directly-Follows Graph (aDFG)} view.~\emph{(b)} shows the \emph{Timeline view with brushed regions} view. Users can brush one or more regions on this graph in order to filter the scope of the analysis~\emph{(b.1}, and~\emph{b.2)}. Two additional controls in~\emph{(c)} show the \emph{activity and path sliders}.} 
	\label{fig:vis-two-brushes}
\end{figure}

The design of the PCE system is shown in Figure~\ref{fig:vis-two-brushes}. It offers an interactive visualisation system with several connected views. The system is implemented using the D3.js JavaScript library and is available as an open-source project.
%\footnote{See \url{https://github.com/yesanton/Process-Change-Exploration-Visualizations}}.

%% file: 5-conclusion.tex
\section{Conclusion}\label{sec:conclusion}
In this paper, we presented the first genuine approach to forecast a process model as a whole. To this end, we developed a technique based on time series analysis of DF relations to forecast entire DFGs from historical event data. 
%In this way, we are able to capture process drifts in an accurate way, as demonstrated in the evaluation using our Process Change Exploration system. 
In this way, we are able to make promising forecasts regarding the future development of the process, including whether process drifts or major changes might occur in particular parts of the process. 
The presented forecasting approach is supported by the Process Change Exploration system, which allows analysts to compare various parts of the past, present, and forecasted future behaviour of the process.
%While further investigation is certainly required, 
Our empirical evaluation demonstrates that, most notably for reduced process models with medium-sized alphabets, we can obtain below 15\% MAPE in terms of conformance to the true models.

In future research, we plan to evaluate the use of machine learning techniques for process model forecasting. More specifically, we aim at using recurrent neural networks or their extension in long short-term memory networks (LSTMs) and transformer-based architectures, as well as hybrid methods or ensemble forecasts with the traditional time series approaches presented here.
Furthermore, we want to explore opportunities for enriching our forecasted process models with confidence intervals by calculating the entropic relevance at different confidence levels and reporting the confidence intervals in the PCE system.
Finally, we will conduct design studies with process analysts to evaluate the usability of different visualisation techniques.